\documentclass[sigconf]{acmart}
\AtBeginDocument{%
  }
\usepackage{multirow}
\usepackage{subcaption}
\acmConference[ISMSI '25]{April 26--28,
  2025}{Tokyo, Japan}

\renewcommand\footnotetextcopyrightpermission[1]{}

\begin{document}

\title{Multi-view Fuzzy Graph Attention Networks for Enhanced Graph Learning}


\author{Jinming Xing}
\affiliation{%
    \institution{North Carolina State University}
    \country{}}
\email{jxing6@ncsu.edu}

\author{Dongwen Luo}
\affiliation{%
    \institution{South China University of Technology}
    \country{}}
\email{976267567ldw@gmail.com}

\author{Qisen Cheng}
\affiliation{%
    \institution{University of Michigan, Ann Arbor}
    \country{}}
\email{qisench@umich.edu}

\author{Chang Xue}
\affiliation{%
    \institution{Yeshiva University}
    \country{}}
\email{cxue@mail.yu.edu}

\author{Ruilin Xing}
\affiliation{%
    \institution{Guangxi University}
    \country{}}
\email{ruilinxing8@gmail.com}

\renewcommand{\shortauthors}{Jinming et al.}

\begin{abstract}
    Fuzzy Graph Attention Network (FGAT), which combines Fuzzy Rough Sets and Graph Attention Networks, has shown promise in tasks requiring robust graph-based learning. However, existing models struggle to effectively capture dependencies from multiple perspectives, limiting their ability to model complex data. To address this gap, we propose the Multi-view Fuzzy Graph Attention Network (MFGAT), a novel framework that constructs and aggregates multi-view information using a specially designed Transformation Block. This block dynamically transforms data from multiple aspects and aggregates the resulting representations via a weighted sum mechanism, enabling comprehensive multi-view modeling. The aggregated information is fed into FGAT to enhance fuzzy graph convolutions. Additionally, we introduce a simple yet effective learnable global pooling mechanism for improved graph-level understanding. Extensive experiments on graph classification tasks demonstrate that MFGAT outperforms state-of-the-art baselines, underscoring its effectiveness and versatility.
\end{abstract}



\maketitle

\section{Introduction}
Fuzzy rough sets have emerged as a powerful mathematical framework for analyzing the fuzzy and vague relationships between objects \cite{dubois90rough,gao22param}. By bridging the concepts of rough sets and fuzzy sets, fuzzy rough sets have been widely applied in various fields, including feature selection \cite{gao22param}, soft computing \cite{zou23novel}, and medical diagnosis \cite{xing22weighted}. These applications leverage the framework's ability to model uncertainty and imprecision effectively. Despite its success, integrating fuzzy systems into the realm of graph-based data analysis remains relatively underexplored.

Graph Neural Networks (GNNs), on the other hand, have become a cornerstone of graph-based machine learning. By encoding graph structures into neural network architectures, GNNs have achieved remarkable success in tasks such as node classification \cite{jin24graphsurvey}, link prediction \cite{xing24enhancing}, and graph classification \cite{luo23towards}. Attention mechanisms \cite{vas17attention,petar17graphattention}, a significant innovation in GNNs, allow these models to focus on the most relevant parts of the graph, thereby improving their ability to handle complex dependencies. However, the fusion of fuzzy systems with GNNs, which could further enhance their ability to manage uncertainty, remains in its infancy.

The introduction of the Fuzzy Graph Attention Network (FGAT) \cite{xing24enhancing} marked an important milestone in integrating fuzzy rough sets with GNNs. FGAT leverages fuzzy lower and upper approximations to compute high-quality negative edges, improving its performance in link prediction tasks \cite{xing24enhancing}. Another extension, FGATT \cite{xing24fgatt}, combines FGAT with Transformer encoders to model spatial-temporal information, demonstrating its utility in wireless data imputation. However, these models have inherent limitations. Specifically, neither FGAT nor FGATT considers the multiple perspective dependencies often hidden within the data. This lack of multi-view modeling can lead to incomplete representations and suboptimal performance.

To address these challenges, we propose the Multi-view Fuzzy Graph Attention Network (MFGAT). The core innovation of MFGAT lies in its Transformation Block, which dynamically captures multi-view information by transforming data from various perspectives. The transformed information is aggregated using a weighted sum mechanism, providing a comprehensive representation that can be effectively processed by FGAT. Additionally, we introduce a learnable global pooling mechanism to generate robust graph-level representations, further enhancing the model's utility in graph classification tasks.

The contributions of this paper can be summarized as follows:
\begin{itemize}
    \item We design a novel Transformation Block to dynamically capture and aggregate multi-view information, addressing the limitations of existing methods.
    \item Based on FGAT, we introduce MFGAT, which combines fuzzy relations and multi-view dependencies with a learnable global pooling mechanism for enhanced graph-level understanding.
    \item Experimental results on graph classification tasks demonstrate the superior performance of MFGAT compared to state-of-the-art baselines, validating the effectiveness of our approach.
\end{itemize}
By bridging the gap between fuzzy systems and graph neural networks, this work advances the field and provides a foundation for further exploration into multi-view graph modeling.

\section{Related Work}
In this section, we provide an overview of fuzzy rough sets, graph neural networks, and multi-view learning, highlighting their significance in modern machine learning. We also discuss the limitations of existing approaches and motivate the need for integrating these concepts in our proposed framework.

\subsection{Fuzzy Rough Sets}
Fuzzy rough sets are a powerful mathematical tool that combines the strengths of rough sets and fuzzy sets to model uncertainty and imprecision in data \cite{ji21fuzzy,xing22weighted}. Introduced by Dubois and Prade in the early 1990s, fuzzy rough sets have found widespread applications in various domains, including feature selection, classification, and decision-making \cite{gao22param,xing22weighted,ye23decision}. The core idea behind fuzzy rough sets is to define lower and upper approximations using fuzzy relations \cite{xing22weighted}, which provide a robust framework for analyzing vague and imprecise data.

Numerous studies have leveraged fuzzy rough sets in real-world applications \cite{yuan21attribute}. For instance, Gao et al. \cite{gao22param} proposed a novel attribute reduction framework based on fuzzy rough sets by maximizing parameterized entropy. More recently, fuzzy rough sets have been applied in medical diagnosis \cite{xing22weighted}, where Xing et al. designed a disease classification framework especially for noisy and semi-supervised learning scenario. Specifically, a novel noisy data filter called bad-point was proposed and a fuzzy rough sets-based weighted tri-training was designed for unlabeled data. Despite these advances, the integration of fuzzy rough sets with modern machine learning techniques, such as GNNs, remains a relatively unexplored area.

\subsection{Graph Neural Networks}
Deep Learning models \cite{xiao23context,xiao23corporate,mai24financial,yangbotnet,linresearch,yangrac} have found innovative applications across various fields like image analysis \cite{zhaoresearch, yangcnn}, virtual reality \cite{yangzhan}, sequences modeling \cite{dengresearch}, and emotion recognition \cite{yangarxiv1}. Among them, Graph Neural Networks have emerged as a dominant paradigm for learning on graph-structured data \cite{petar23everything}. By generalizing neural network architectures to graphs, GNNs have achieved state-of-the-art performance in tasks such as node classification, link prediction, and graph classification \cite{xing24enhancing,wu20graphsurvey,jin24graphsurvey,xing24fgatt,xu23comprehensive}. Various GNN architectures, such as Graph Convolutional Networks (GCNs) \cite{chen20simple}, Graph Attention Networks (GATs) \cite{petar17graphattention}, and Graph Isomorphism Networks (GINs) \cite{xu18gin}, have been proposed to capture the structural and semantic properties of graphs effectively. Among these, GATs have gained significant attention due to their ability to assign varying importance to different nodes and edges through attention mechanisms, enabling more expressive graph representations \cite{brody21attentive}.

In recent years, there has been increasing interest in extending GNNs to handle fuzzy or uncertain relationships. Notably, Fuzzy Graph Attention Networks (FGATs) \cite{xing24enhancing} have been developed to incorporate fuzzy rough set theory into graph learning. By utilizing fuzzy lower and upper approximations, FGAT enhances the representation of uncertain relationships in graphs, improving tasks like link prediction. However, existing approaches like FGAT and its extension FGATT \cite{xing24fgatt}, which integrates Transformer encoders for spatial-temporal modeling, do not account for multi-view dependencies. This limitation reduces their ability to model data comprehensively, motivating the need for approaches that capture diverse perspectives in graph-based learning.

\subsection{Multi-view Learning}
Multi-view learning has gained significant attention for its ability to integrate information from diverse perspectives to improve learning performance \cite{xing22weighted,yan21multiview}. In graph-based tasks, multi-view approaches aim to capture dependencies that might be overlooked when considering a single perspective \cite{wang19graphmultiview}. Techniques like weighted sum mechanisms \cite{xing24pooling} and attention-based aggregation \cite{brody21attentive} have been employed to combine multiple views effectively. However, integrating multi-view learning with fuzzy systems in GNNs remains underexplored.

Our proposed Multi-view Fuzzy Graph Attention Network (MFGAT) bridges this gap by introducing a Transformation Block that dynamically captures and aggregates multi-view dependencies. By combining fuzzy rough set theory with robust multi-view modeling, MFGAT represents a significant step forward in the field, addressing the limitations of existing GNN architectures and fuzzy graph models.

\section{Methodology}
In this section, we introduce the core components of the proposed Multi-view Fuzzy Graph Attention Network (MFGAT). First, we describe the Transformation Block for learning multi-view dependencies. Next, we discuss the FGAT convolution layers, followed by the learnable global pooling mechanism for graph-level understanding. Finally, we summarize the overall framework.
\begin{figure}
    \centering
    \includegraphics[width=0.99\linewidth]{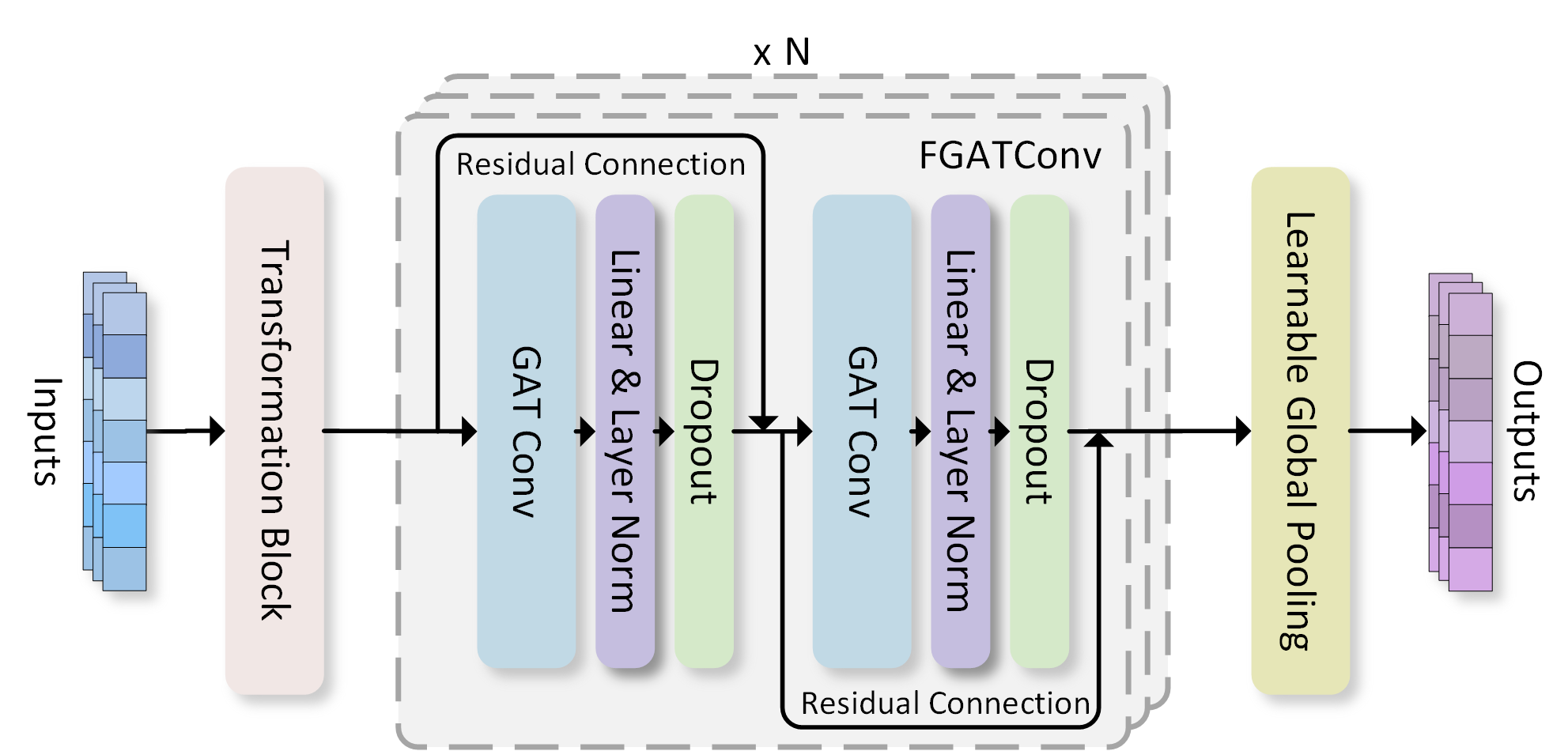}
    \caption{Multi-view Fuzzy Graph Attention Network}
    \label{fig:MFGAT}
\end{figure}

\subsection{Transformation Block}
Given a graph with node information $X = \{x_1, x_2, \cdots, x_n\}$, where $x_i \in \mathbb{R}^d$, $n$ is the number of nodes in the graph, and $d$ is the feature dimension, the transformed feature of node $i$ in view $j$ is defined as:
\begin{equation}
    x_i^j=W_jx_i + b_j
    \label{eq:transformation}
\end{equation}
where $W_j$ and $b_j$ are the learnable weight and bias parameters for view $j$.

Using Equation \ref{eq:transformation}, we compute the multi-view representation of node $i$ as $x_i^{mv} = \{x_i^1, x_i^2, \cdots, x_i^m\}$, where $m$ is the number of specified views. Aggregating these multi-view representations into a unified representation is a critical step. As stated in \cite{xing24pooling}, Mean, Max, and Weighted Sum are three common pooling techniques. Each pooling method has unique strengths: Mean pooling provides stability and robustness, Max pooling emphasizes salient features, and Weighted Sum pooling offers flexibility but requires careful optimization \cite{xing24pooling}.

In this paper, we adopt the Weighted Sum pooling mechanism for its versatility in multi-view scenarios. The unified multi-view representation of node $i$ is defined as:
\begin{equation}
    x_i^{uni}=W \odot x_i^{mv}
\end{equation}
where $W = \{w_1, w_2, \cdots, w_m\}$ represents the learnable weights, and $\odot$ denotes element-wise multiplication.

The Transformation Block is applied to all nodes, producing their unified representations, $X^{uni} = \{x_1^{uni}, x_2^{uni}, \cdots, x_n^{uni}\}$. These representations, along with the adjacency matrix $E$, are input to the FGAT layers for robust spatial dependency modeling.

\subsection{FGAT Convolution}
An FGAT layer \cite{xing24enhancing} consists of two identical sub-modules, each comprising a GAT Convolution, a Linear Layer, Layer Normalization, a Residual Connection, and a Dropout Layer. Below, we briefly introduce each component:
\begin{itemize}
    \item \textbf{GAT Convolution}:
          Graph Attention Network (GAT) employs an attention mechanism to compute the importance of edges between nodes, enabling dynamic aggregation of neighboring node features. The attention coefficient $\alpha_{ij}$ is calculated as:
          \begin{equation}
              \alpha_{ij} = \text{softmax}_j\left(\text{LeakyReLU}\left(a^T[W x_i || W x_j]\right)\right)
          \end{equation}
          where $W$ is a learnable weight matrix, $a$ is a learnable attention vector, and $||$ denotes concatenation.

    \item \textbf{Layer Normalization}:
          Layer normalization stabilizes training by normalizing the summed inputs to a layer. It is defined as:
          \begin{equation}
              x^i=\frac{x_i - \mu}{\sigma + \epsilon} \cdot \gamma + \beta
          \end{equation}
          where $\mu$ and $\sigma$ are the mean and standard deviation of the input features, and $\gamma, \beta$ are learnable parameters.

    \item \textbf{Residual Connection}:
          Residual connections enable deeper networks by adding the input to the output of a layer, defined as:
          \begin{equation}
              y_i = f(x_i) + x_i
          \end{equation}
          where $f(x_i)$ is the transformation applied to the input $x_i$.

    \item \textbf{Dropout Layer}:
          Dropout introduces stochasticity during training to prevent overfitting. A fraction $p$ of the neurons is randomly set to zero:
          \begin{equation}
              \hat{x}_i={\begin{cases}  x_i, & \text{with probability } 1 - p \\  0, & \text{with probability } p  \end{cases}}
          \end{equation}
\end{itemize}

By stacking multiple FGAT layers, the model captures multi-hop information, which is subsequently utilized in the learnable global pooling mechanism for graph-level understanding.

\subsection{Learnable Global Pooling}
Pooling mechanisms play a critical role in aggregating node-level representations into a graph-level representation. As noted in \cite{xing24pooling}, different pooling techniques cater to specific needs:
\begin{itemize}

    \item \textbf{Mean Pooling}: Efficient for computational constraints.
    \item \textbf{Max Pooling}: Effective for tasks requiring maximum positive detection.
    \item \textbf{Weighted Sum Pooling}: Suitable for flexible, adaptable applications.
\end{itemize}

Given the multi-purpose nature of our framework, we adopt the Weighted Sum pooling mechanism. However, to fully capture the multi-view dependencies, we extend it to a multi-view Weighted Sum mechanism, defined as:
\begin{equation}
    Pool_{ws}^{mv}(X) = W^{mv} \odot \left(||^j W^j X \right)
\end{equation}
where $W^jX$ represents the weighted sum for view $j$, $||$ denotes concatenation across views, $\odot$ indicates element-wise multiplication, and $W^{mv}, W^j$ are learnable parameters.

This multi-view Weighted Sum mechanism aggregates the weighted sums of different views and combines them to generate a unified graph representation. This approach ensures the comprehensive capture of multi-view dependencies.

\subsection{Multi-view Fuzzy Graph Attention Network}
As illustrated in Figure \ref{fig:MFGAT}, we summarize the proposed Multi-view Fuzzy Graph Attention Network (MFGAT) as follows:
\begin{itemize}
    \item \textbf{Transformation Block}: Learns multi-view dependencies by transforming node features into view-specific representations and aggregating them into a unified representation using Weighted Sum pooling.
    \item \textbf{FGAT Layers}: Model spatial dependencies using fuzzy graph convolutions, supported by attention mechanisms and normalization techniques.
    \item \textbf{Learnable Global Pooling}: Aggregates node-level features into a graph-level representation using the multi-view Weighted Sum mechanism.
\end{itemize}

By combining these components, MFGAT achieves robust graph-based learning, capturing both fuzzy relations and multi-view dependencies for enhanced performance in graph-level tasks.

\section{Experiments}
In this section, we first introduce the datasets used for performance evaluation. Following that, we elaborate on the experimental settings and baselines. Finally, we present and discuss the experimental results.

\subsection{Datasets}
We selected three commonly used graph-classification datasets from TUDataset \cite{morris20tudatasets} for performance evaluation, as summarized in Table \ref{tab:datasets summary}. Specifically, the PROTEINS dataset contains 1113 graphs, while the other two datasets consist of more than 4000 graphs each. All three are binary classification datasets with undirected links and multi-dimensional features per node. Detailed information about these datasets is provided in Table \ref{tab:datasets summary}.
\begin{table}[htbp]
    \centering
    \caption{Datasets Summary}
    \begin{tabular}{lccc}
        \toprule
                            & \textbf{PROTEINS} & \textbf{NCI1} & \textbf{Mutagenicity} \\
        \midrule
        \textbf{\#Graphs}   & 1113              & 4110          & 4337                  \\
        \textbf{\#Classes}  & 2                 & 2             & 2                     \\
        \textbf{Avg. Nodes} & 39.06             & 29.87         & 30.32                 \\
        \textbf{Avg. Edges} & 72.82             & 32.30         & 30.77                 \\
        \textbf{\#Features} & 3                 & 37            & 14                    \\
        \bottomrule
    \end{tabular}%
    \label{tab:datasets summary}%
\end{table}%

\subsection{Experimental Settings and Baselines}
Each dataset was split into training, validation, and test sets, occupying 70\%, 10\%, and 20\% of the total data, respectively. The maximum number of training epochs was set to 200. We used Adam as the optimizer, with a learning rate of 0.01. Cross-validation and early stopping were employed for hyperparameter tuning and model selection. Accuracy was chosen as the evaluation metric.

Our model was compared against several state-of-the-art frameworks, including:
\begin{itemize}

    \item \textbf{GCN \cite{chen20simple}}: Uses spectral convolution to aggregate node information.
    \item \textbf{GAT \cite{petar17graphattention}}:  Based on attention mechanisms for weighted neighbor aggregation.
    \item \textbf{GraphSAGE \cite{ham17graphsage}}: Samples and aggregates features from neighboring nodes for scalable learning.
    \item \textbf{FGAT \cite{xing24enhancing}}: Integrates Fuzzy Rough Sets into Graph Attention Networks to improve graph-based learning.
\end{itemize}

\subsection{Results}
As presented in Table \ref{tab:experiment results}, MFGAT, with the number of views set to 3, outperformed all baselines across the three datasets. Specifically, GCN exhibited the weakest performance on the PROTEINS dataset but achieved the second-best performance on the Mutagenicity dataset. GraphSAGE, recognized for its versatility as a GNN template, achieved intermediate results but outperformed GAT in all three datasets. FGAT, the fuzzy extension of GAT, surpassed GAT in both the NCI1 and Mutagenicity datasets and outperformed GraphSAGE in the NCI1 dataset, demonstrating the value of incorporating fuzzy relations. Notably, our proposed model, MFGAT, which incorporates multi-view fuzzy dependencies, consistently outperformed all baselines, confirming its superior performance.
\begin{table}[htbp]
    \centering
    \caption{Performance Evaluation on Graph Classification Task}
    \begin{tabular}{lccc}
        \toprule
                           & \textbf{PROTEINS} & \textbf{NCI1}   & \textbf{Mutagenicity} \\
        \midrule
        \textbf{GCN}       & 0.6951            & 0.6488          & 0.7604                \\
        \textbf{GAT}       & 0.7220            & 0.6513          & 0.7189                \\
        \textbf{GraphSAGE} & 0.7354            & 0.6615          & 0.7546                \\
        \textbf{FGAT}      & 0.7188            & 0.6712          & 0.7228                \\
        \textbf{MFGAT}     & \textbf{0.7630}   & \textbf{0.6865} & \textbf{0.8097}       \\
        \bottomrule
    \end{tabular}%
    \label{tab:experiment results}%
\end{table}%

\begin{figure}[htbp]
    \centering
    \includegraphics[width=0.75\linewidth]{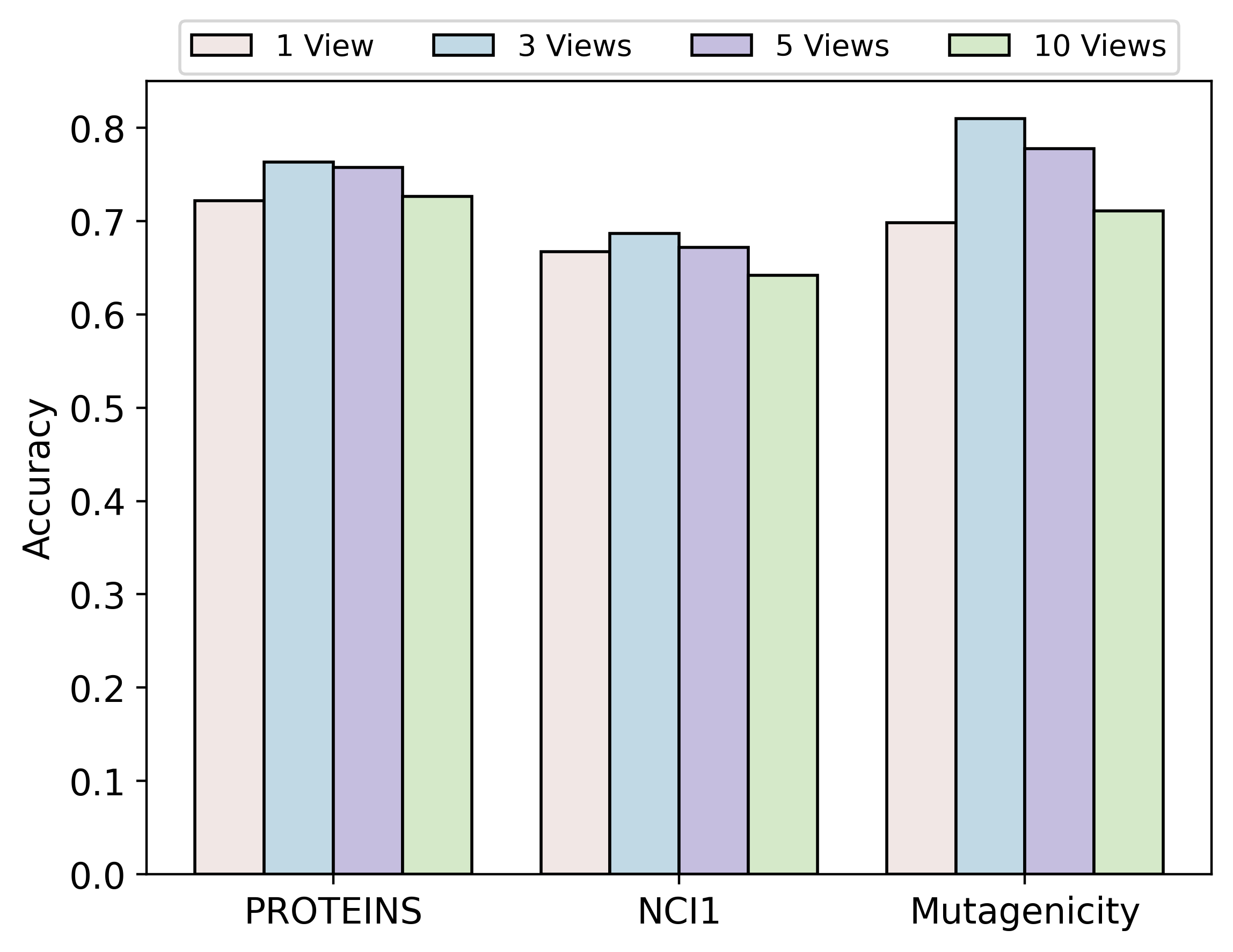}
    \caption{Impact of Number of Views on MFGAT}
    \label{fig:views}
\end{figure}

To further evaluate the impact of the number of views on MFGAT, we conducted experiments where the number of views was set to 1, 3, 5, and 10. The results are illustrated in Figure \ref{fig:views}.

A consistent trend can be observed across all three datasets. MFGAT with three views achieved the best performance, followed by five views. Notably, setting the number of views too low or too high may adversely affected performance. Specifically, too few views limit the model's ability to capture diverse perspectives while too many views may introduce more noise than information, increase the risk of overfitting, and require higher computational resources. In practice, the optimal number of views should be selected based on specific task requirements and computational constraints to achieve the best balance between performance and efficiency.

\section{Conclusion}
In this paper, we presented the Multi-view Fuzzy Graph Attention Network (MFGAT), a novel framework designed to overcome the limitations of existing Fuzzy Graph Attention Networks by incorporating multi-view dependencies. By introducing the Transformation Block, we enable the dynamic modeling of information from multiple perspectives, which is critical for capturing the complex relationships inherent in real-world data. Furthermore, the proposed learnable global pooling mechanism ensures robust graph-level representations, further enhancing the model's performance. Through extensive experiments on graph classification tasks, MFGAT demonstrated superior performance compared to state-of-the-art methods, validating the effectiveness of our approach. These results highlight the importance of multi-view modeling in graph neural networks and open avenues for further research into integrating fuzzy systems with advanced graph learning architectures. Future work could explore the extension of MFGAT to other graph-based tasks, such as node classification and link prediction, and investigate its applicability to diverse real-world scenarios.


\end{document}